\documentclass[10pt]{article} 
\usepackage[accepted]{rlc}

\usepackage{amssymb}            
\usepackage{mathtools}          
\usepackage{mathrsfs}           
\mathtoolsset{showonlyrefs}     
\usepackage{graphicx}           
\usepackage{subcaption}         
\usepackage[space]{grffile}     
\usepackage{url}                
\usepackage{wrapfig}

\usepackage{hyperref}
\usepackage{array}
\usepackage{float}
\usepackage{algorithm}
\usepackage{algpseudocode}

\usepackage{color}

\title{Representation Alignment from Human Feedback for Cross-Embodiment Reward Learning from Mixed-Quality Demonstrations}

\author{Connor Mattson$^*$ \\
    c.mattson@utah.edu\\
    Kalhert School of Computing\\
    University of Utah
    \And
    Anurag Aribandi$^*$  \\
    anurag.aribandi@utah.edu \\
    Kalhert School of Computing\\
    University of Utah
    \And
    Daniel S. Brown \\
    daniel.s.brown@utah.edu\\
    Kalhert School of Computing\\
    University of Utah
    }

\begin{document}

\maketitle
\def\thefootnote{$*$}\footnotetext{Equal contribution.}\def\thefootnote{\arabic{footnote}}
\def\thefootnote{}
\footnotetext{Videos and code available at \url{https://sites.google.com/view/cross-irl-mqme/home}}\def\thefootnote{\arabic{footnote}}

\begin{abstract}
We study the problem of cross-embodiment inverse reinforcement learning, where we wish to learn a reward function from video demonstrations in one or more embodiments and then transfer the learned reward to a different embodiment (e.g., different action space, dynamics, size, shape, etc.). Learning reward functions that transfer across embodiments is important in settings such as teaching a robot a policy via human video demonstrations or teaching a robot to imitate a policy from another robot with a different embodiment. However, prior work has only focused on cases where near-optimal demonstrations are available, which is often difficult to ensure. By contrast, we study the setting of cross-embodiment reward learning from mixed-quality demonstrations. We demonstrate that prior work struggles to learn generalizable reward representations when learning from mixed-quality data. We then analyze several techniques that leverage human feedback for representation learning and alignment to enable effective cross-embodiment learning. Our results give insight into how different representation learning techniques lead to qualitatively different reward shaping behaviors and the importance of human feedback when learning from mixed-quality, mixed-embodiment data.
\end{abstract}

\section{INTRODUCTION}
Inverse reinforcement learning (IRL)~\citep{arora2021survey} seeks to learn a reward function from observed agent behavior. However, the field of imitation learning~\citep{hussein2017imitation} has developed numerous techniques for direct policy learning from observed agent behavior. So why learn a reward function? From the earliest days of IRL research, \cite{ng2000algorithms} and others have argued that reward functions provide a succinct description of behavior. 
Indeed, \cite{ng2000algorithms} notes that the field of reinforcement learning (RL) is based on the idea that the reward function is ``the most succinct, robust, transferable definition of a task.'' 
Thus, if we can learn a reward function from observing an agent in one task, it should be the case that we can use that reward function to help teach the same task to agents with different embodiments (e.g., different action space, dynamics, size and shape, etc.). The idea of allowing agents with different embodiments to learn from each other is typically called \textit{cross-embodiment}~\citep{zakka2022xirl} or \textit{cross-domain}~\citep{niu2024comprehensive} learning. In this paper we focus on cross-embodiment IRL, with the goal of learning robust reward functions that can transfer across different embodiments.

Cross-embodiment reward learning would enable robots to learn rewards from watching humans perform tasks and would allow robots and other AI agents to learn by watching other agents. 
However, given an embodiment mismatch between the demonstrator and the learner, we cannot simply imitate the actions of the demonstrator since the action spaces may be completely different. 
Furthermore, the actions are typically unavailable when learning only from video observations~\citep{torabi2019recent}. 
Learning an embodiment-independent reward function is a compelling solution to the problem of cross-embodiment policy learning as it should allow an agent of any embodiment to learn how to perform the desired task via reinforcement learning.  
However, prior work has shown that reward functions learned from demonstrations are often entangled with the dynamics, making cross embodiment transfer difficult~\citep{fu2017learning}. 

Recently, \cite{zakka2022xirl} developed XIRL, a novel approach for cross-embodiment reward learning from video demonstrations. Using near-optimal demonstrations across several different embodiments they first learn an embodiment-invariant representation using temporal cycle-consistency~\citep{dwibedi2019temporal}. The base assumption this method uses is that there is a temporal similarity between the data it is trained on, i.e., there are similar frames or checkpoints that each video demonstration will share with another.  The reward is then formulated as the distance between the current state embedding to that of a goal embedding and this learned reward is optimized via RL to achieve generalization to an unseen embodiment. However, one of the main limitations of this recent breakthrough by \cite{zakka2022xirl} is that, in order to ensure temporal similarity when performing representation learning, the approach requires \textit{near-optimal demonstrations across each embodiment}. 
Prior work has shown that human demonstrations and other interactions with AI systems can be noisy~\citep{chuck2017statistical,mandlekar2022matters}, irrational~\citep{chan2021human,ghosal2023effect}, and sometimes adversarial~\citep{wolf2017we,jagielski2018manipulating,oravec2023rage}. Thus, assuming that demonstration data is near-optimal is unlikely to be true in practice.

We study several different approaches for performing cross-embodiment reward learning from \textit{mixed-quality demonstrations}: (1) \textit{Cross-Embodiment Reinforcement Learning from Human Feedback} where we learn a reward function end-to-end from preferences over demonstrations from different embodiments in our training dataset and use this learned reward function to perform reinforcement learning~\citep{christiano2017deep}; (2) \textit{Cross-Embodiment Representation Learning from Preferences}: We explore techniques that seek to use human preference labels to learn a state representations~\citep{tian2023matters} and then formulate the reward as the distance between the learned state embeddings and a goal embedding; and
(3) \textit{XIRL-Buckets}: A method that seeks to apply a temporal cycle-consistency representation learning to mixed-quality, mixed-embodiment data by leveraging high-level knowledge of the relative goodness of demonstrations by first binning the demonstrations into several "buckets" or groups based on ordinal labels denoting their goodness and then performing temporal cycle-consistency representation learning within each bucket. 






The primary contributions of this work are:
    (1) We propose and formalize the new problem of cross-embodiment reward learning from mixed-quality data;
    (2) We study a range of algorithmic approaches for this problem setting that build on and combine ideas from representation learning and alignment from human feedback;
    (3) We empirically study the RL performance, learned rewards and learned representations when learning with mixed data and show that prior approaches fail to perform well in this setting;
    (4) We provide empirical evidence that approaches that leverage human feedback information about the relative quality of the data are often able to learn transferable representations and corresponding rewards that transfer across embodiments even when learning from mixed-quality demonstrations. However, these methods still fail to achieve the performance of methods that learn only from high-quality demonstrations, showing that there is a need for further research into how to best learn from mixed-quality, mixed embodiment data. 

\section{PRIOR WORK}

\textbf{Imitation Learning from Observation}
Our work seeks to leverage video observations from one embodiment and learn to transfer these policies to new embodiments. Thus, our work falls within the general area of imitation learning from observation~\citep{torabi2019recent}. Early work on inverse reinforcement learning learned reward functions based on state observations of demonstrations~\citep{abbeel2004apprenticeship,ziebart2008maximum}. Other work proposed imitation learning from state observations via model-based behavioral cloning~\citep{torabi2018behavioral} by learning an inverse dynamics model and subsequent work proposed to explicitly minimize inverse dynamics disagreement~\citep{yang2019imitation}. However, these prior works focused on learning from low-dimensional state observation sequences. More recent work studies cases where the goal is to learn from video observations~\citep{torabi2018generative,liu2018imitation,salimans2018learning,goo2019one,kidambi2021mobile}.

\textbf{Cross-Embodiment Reward and Policy Learning}
Much prior work has focused on cross-embodiment learning~\citep{niu2024comprehensive} when given full access to the demonstrating agent's statespace.
\cite{fu2017learning} demonstrate that learned reward functions are often entangled with embodiment dynamics and propose an adversarial reward learning approach that seeks to learn an unshaped, state-only reward and
\cite{fickinger2021cross} perform cross-domain imitation learning based on optimal transport but both methods are restricted to low-dimensional state spaces rather than video observations.
Recently, researchers have proposed methods that can scale to settings with partial observability, where only video observations of demonstrations are available.
\cite{zakka2022xirl} propose a method for cross-embodiment inverse reinforcement learning from videos by finding correspondences across time across multiple videos. 
\cite{xu2023xskill} consider an alternative approach that leverages optimal human and robot demonstrations to learn resuable skills from videos. \cite{tian2023matters} propose using human triplet preferences to align representations in one embodiment and show that these representations can transfer to new embodiments.
Other work seeks to train robot foundation models that can then be fine-tuned on arbitrary robot embodiments~\citep{padalkar2023open}.
Most prior work on imitation learning from observations and cross-embodiment reward and policy learning focuses on learning from near-optimal demonstrations. By contrast, we seek to perform cross-embodiment learning from mixed-quality data.

\textbf{Learning from Suboptimal Demonstrations}:
When ground-truth rewards are known, it is common to initialize a policy using demonstrations and then improve this policy using reinforcement learning \citep{hester2018deep,gao2018reinforcement,wilcox2022monte}. However, these methods typically do not consider embodiment mismatches and rely on designing a reward design which can easily lead to unintended behaviors \citep{ng1999policy,amodei2016concrete,booth2023perils}. Other work learns from demonstrations that are labeled good or bad~\citep{grollman2011donut,shiarlis2016inverse} or are robust to a small number of unlabeled, poor demonstrations~\citep{zheng2014robust,choi2019robust}. However, these prior works do not consider embodiment mismatch and require low-dimensional state observations.
Prior work has considered learning reward functions from pairwise preferences over trajectories~\citep{wirth2017survey,christiano2017deep} and has shown the ability of these methods to extrapolate beyond the performance of suboptimal demonstrations~\citep{brown2019extrapolating,brown2020better}. While there has been substantial progress recently in learning from pairwise preferences~\citep{lee2021pebble,park2021surf,bobu2023sirl,rafailov2024direct,myers2022learning,shin2023benchmarks,liu2023efficient,wilde2021learning,ouyang2022training,karimi2024reward}, prior work does not consider the cross-embodiment setting that we explore in this paper.

\section{PROBLEM FORMULATION}
Our problem setting is inspired by~\cite{zakka2022xirl}, who consider cross-embodiment IRL. However, in contrast to \cite{zakka2022xirl}, we seek to learn from mixed-quality, mixed-embodiment data.
We investigate the problem of learning an agent-agnostic representation of a task, $T$, given a dataset of videos depicting agents performing the task. Formally, we define a dataset $D$ as a collection of state-only video demonstrations $D = \{v_0, v_1, \dots, v_i\}$, where each video contains a sequence of frames (2D images), $v_i = \{v_i^0, v_i^1, \cdots, v_i^j\}$ that depict the agent executing the task.
In contrast to prior work~\citep{zakka2022xirl}, we consider a set of \textit{mixed-quality} demonstrations, where it is no longer guaranteed that agents will reach the goal state at the end of every demonstration. 
We refer to the demonstration set as \textit{mixed-embodiment} if it contains demonstrations from more than one agent embodiment performing the same task. 

\textbf{Problem Statement:} Given a task, $T$, and a mixed-quality, mixed-embodiment (MQME) demonstration dataset, $D$,  can we learn an embodiment-agnostic approximation, $\hat{r}$, of the ground truth reward function? Furthermore, can this learned reward function be used to successfully accomplish task $T$ by performing RL on $\hat{r}$ with an unseen embodiment?

\section{PRELIMINARIES}\label{sec:prelims}
In this section we describe in detail the XIRL algorithm, an unsupervised method of learning embodiment-agnostic representations of tasks proposed by \cite{zakka2022xirl}. XIRL assumes access to a video dataset, $D = \{v_0, v_1, \dots, v_i\}$, of successful video demonstrations, $v_i$, for a task. Each video demonstration is an observation-only video (demonstrator actions are unobserved) containing a sequence of video frames, $v_i = \{v_i^1, v_i^2, \cdots, v_i^j\}$. XIRL assumes that $D$ contains video demonstrations from multiple agent embodiments, all performing the same task. 

To learn from multi-embodiment data, \cite{zakka2022xirl} seek to learn a useful representation for cross-embodiment learning that aligns task progress in one embodiment to task progress in a different embodiment. 
To address this, XIRL relies on  temporal cycle-consistency (TCC)~\citep{dwibedi2019temporal} learning to establish task-aligned feature representations of cross embodiment demonstrations that captures information about the task itself, rather than the agent executing the task. 
The goal of TCC is to train an encoder, $\phi$, that takes as input a video frame image, $v^s$, corresponding to state $s$, and outputs an embedding vector $\phi(v^s)$.
One of the primary benefits of XIRL's use of TCC is that it does not require embodiment labels and is completely unsupervised. First, random mini-batches of video trajectories are sampled from $D$. Given $\phi$, each video trajectory can be represented as a sequence of embedded images, $V_i = \{\phi(v_i^1), \phi(v_i^2), \cdots, \phi(v_i^{L_i})\}$, where $L_i = |v_i|$. Given a mini-batch of embedded videos, $\phi$ is updated by taking pairs of sequences $V_i$ and $V_j$ and computing a TCC Loss which aligns a random frame index, $t$, in $V_j$ to the corresponding soft-nearest-neighbor frame, $t'$, in $V_i$ by minimizing the mean-squared error between the frame indices, $L_{ij}^t = (t'-t)^2$~\citep{zakka2022xirl}. 

Following the self-supervised training of the encoder $\phi$, XIRL grounds an embodiment-agnostic reward function to the demonstration set by computing a \textit{goal embedding}. Because XIRL assumes that the demonstrations provided are always near-optimal, the last frame of every sequence, $v_i^{L_i}$, can be assumed to represent a state where the agent successfully completed the task. Therefore, the average of these final frames' embeddings are averaged to create a goal state embedding, $g=\frac{1}{N} \sum^N_{i=1} \phi(v_i^{L_i})$.
Finally, both the encoder and the goal embedding are used to provide a reward signal during reinforcement learning. Specifically, the reward is the negative signed distance to goal, 
\begin{equation}
    \label{eq:reward_form}
    r(s) = -\frac{1}{\kappa} \cdot \lVert \phi(s) - g \rVert_2^2,
\end{equation} 
where $\kappa$ is a scaling parameter. 

In the following sections, we build on the foundational work of \cite{zakka2022xirl} to study cross-embodiment learning when multi-embodiment demonstrations are of mixed quality and may not always successful complete the desired task.


\section{METHODS}
We seek to learn reward representations that generalize to unseen agent embodiments,
In contrast to XIRL~\citep{zakka2022xirl}, 
we assume that our dataset is MQME: \textit{mixed-quality}, where the quality of the demonstration with respect to task success varies and also \textit{mixed-embodiment} (i.e. demonstrations will be given by agents with various physical embodiments and action spaces).
Dealing with mixed-quality data poses a challenge for two of the main components of the XIRL pipeline: (1) XIRL pretrains the video frame encoder $\phi$ with TCC which assumes that temporal alignment exists between all demonstrations, e.g., each demonstration starts in a similar start state configuration and is assumed to end at state that corresponds to task success, with several key corresponding intermediate steps. 
However, if some demonstrations are of mixed quality, this temporal alignment may not exist between pairwise samples of videos. (2) The reward formulation for XIRL depends on a reliable goal approximation, $g$, which is the average embedding from the final frame of all videos in the dataset. By removing the guarantee that all demonstrations successfully complete the task, the XIRL reward function may not guide the agent towards task completion as the goal embedding may no longer be reliable.
In the following sections, we address these concerns and propose and discuss different approaches for performing cross-embodiment reward learning from MQME data.



\subsection{Cross-Embodiment Reinforcement Learning From Human Feedback (X-RLHF)} \label{sec:RLHF}

The simplest way to address the issues of mixed-quality data in XIRL is to try learning a reward end-to-end with using reinforcement learning from human feedback (RLHF)~\citep{christiano2017deep,ouyang2022training}, by having a human provide preference labels over an offline dataset of trajectories~\citep{shin2023benchmarks}. 
In this approach, the demonstration dataset is augmented by a set of pairwise preference labels over the data. For a pair of demonstrations, $(v_i, v_j)$, the notation $v_i \succ v_j$  indicates a preference of demonstration $j$ over demonstration $i$. 
The final form of the data is the triple $(v_i, v_j, \mu)$, where $\mu \in \{(1,0), (0, 1)\}$ represents the human's preference label. For our problem, it is worth noting that the preferences include mixed-embodiment preferences, i.e., $v_i$ may be demonstrated by one embodiment and $v_j$ may be from a different embodiment.

Using these labels, we follow prior work by employing a deep neural network, namely a reward predictor, $\hat{r}$, that maps video frames into a single real-valued reward that can be trained via backpropagation using the standard Bradley-Terry~\citep{bradley1952rank} and Luce-Shepperd~\citep{luce2005individual, shepard1957stimulus} model,
\begin{equation}
    P(v_i \succ v_j) = \frac{\exp \ {\sum\limits_{s \in v_i} \hat{r}(s)}}{\exp \sum\limits_{s \in v_i} \hat{r}(s) + \exp\ {\sum\limits_{s \in v_j} \hat{r}(s)}}
\end{equation}
where $P$ is the softmax probability that $v_i \succ v_j$ based on $\hat{r}$. The learned reward function, $\hat{r}$, is then optimized using a Cross Entropy Loss between the predicted value of $P$ and the preference labels:
\begin{equation}
    \mathcal{L}(\hat{r}) = - \sum\limits_{(v_i, v_j, \mu) \in D} \mu_1 \log P(v_i \succ v_j) + \mu_2 \log P(v_i \succ v_j)
    \label{eq:cross-entropy-prefs-loss} \;.
\end{equation}





While prior work has focused on single-embodiment RLHF, we seek to study cross-embodiment RLHF (X-RLHF). The benefit of using X-RLHF is that it directly learns the reward end-to-end, with no intermediate latent representation or goal embedding used to manually compute the reward. It is a natural choice for MQME data since the only requirement is to have preference labels over trajectories and these trajectories do not need to be optimal nor come from the same embodiment. Although X-RLHF requires more human burden compared to XIRL, using human supervision can lead to better and more human-aligned representations~\citep{bobu2023sirl,mattson2023leveraging,tian2023matters} than purely unsupervised representation learning approaches.



\subsection{Representation Learning from Preferences}\label{sec:representations_from_prefs}
An alternative approach to learning a reward function end-to-end from preferences is to use human feedback to explicitly learn the representation $\phi(s)$~\citep{bobu2023sirl,mattson2023leveraging,tian2023matters}. When provided with MQME data, we hypothesize that representation learning via TCC will fail to learn a correct embedding because both videos may not share the same task-relevant keyframes.
Thus, we propose the use of preferences to learn a better latent embedding that can be used to guide RL via the same reward function used in XIRL (Equation~\eqref{eq:reward_form}). However, Equation~\eqref{eq:reward_form} requires a known or calculated goal embedding, $g$. 
We assume that in addition to the MQME dataset, we have direct and privileged access to a known set of goal states, $G^* = \{g^*_1, g^*_2, \cdots, g^*_N\}$. Note that these could be supplied by the user as a set of states, disjoint from any actual trajectories. Alternatively, these goal states can come from suboptimal trajectories that eventually reach the goal state. Our goal embedding is simply the average embedding over all goals in $G^*$, resulting in 
    $g = \frac{1}{|G^*|}\sum\limits_{g^*_i\in G^*}\phi(g^*_i)$.

One deceptively simple approach is to combine preference learning with the inductive bias in XIRL by using the same underlying architecture and reward function as XIRL, but with supervised learning from pairwise preferences. We call this approach Cross Preference Learning (XPrefs). To do this, we could use the preference data to optimize the representation $\phi(s)$ by maximizing the likelihood of the preference labels, where
\begin{equation} \label{eq:prob-xprefs}
    P(v_i \succ v_j) = \frac{\exp \ {\sum\limits_{s \in v_i} -\lVert \phi(s) - g \rVert_2^2}}{\exp \sum\limits_{s \in v_i} -\lVert \phi(s) - g \rVert_2^2 + \exp\ {\sum\limits_{s \in v_j} -\lVert \phi(s) - g \rVert_2^2}} \; .    
\end{equation}
However, we now have a non-stationary goal embedding, resulting in an ``chicken-and-egg'' cyclic dependency where both the embedding of the demonstration, $\phi(s)$, and $g$, which is a function of $\phi$, will change every time the model updates. 
In Appendix \ref{app:static-vs-dynamic} we explore 
both dynamic and static goal representations in Appendix \ref{app:static-vs-dynamic} and show that a static goal representation results in the best performance for reward learning. However, upon closer inspection, it can be seen that XPrefs is nearly identical to X-RLHF. Indeed, the effect of the goal embedding is lost when $\phi(s) = \phi'(s) + g$, where $\phi'(s)$ is an arbitrary function of state.  Thus, XPrefs is simply X-RLHF with a non-positive reward function. Indeed, in Appendix~\ref{app:static-vs-dynamic} we empirically compare the performance of X-RLHF and XPrefs and find they are nearly identical. 

The crux of the problem with XPrefs is that we are still trying to learn the reward function and representation simulataneously. Instead, motivated by recent work~\citep{tian2023matters}, we seek to first learn an aligned representation using human feedback and then use this fixed representation as the representation $\phi$ using the same reward function as XIRL (Equation~\eqref{eq:reward_form}). This eliminates the chicken-and-egg problem since the goal embedding is not calculated until after the representation is learned. \cite{tian2023matters} propose the use of triplet preference queries as a way to learn an aligned representation. We follow their approach and seek a representation that is aligned with human's preferences. We obtain ranked triplets over MQME data $v_i \succ v_j \succ v_k$ where we assume rankings are based on the human's internal reward function. We then learn a representation $\phi$ using the Bradley-Terry model~\cite{bradley1952rank} 
\begin{equation}
    P(v_i \succ v_j \succ v_k) = \frac{\exp -d(\phi(v_i), \phi(v_j))}{\exp-d(\phi(v_i), \phi(v_j)) + \exp -d(\phi(v_j), \phi(v_k))}
\end{equation}
where $d$ is a distance metric and where we treat $v_i$ as an anchor and $v_j$ as a positive and $v_k$ as a negative in a contrastive loss. Given triplet preferences we can directly backpropagate into the representation $\phi$ to maximize the likelihood of the preference labels.

We call this approach Cross-Embodiment Triplet Representation Learning (XTriplets). We note some differences between our work and \cite{tian2023matters}. \cite{tian2023matters} use a differentiable optimal transport-based distance metric $d$, but do not investigate using a simple distance metric such as the L2 norm that was proposed by \cite{zakka2022xirl}. \cite{tian2023matters} also do not consider training across multiple embodiments, where as we learn from MQME data. To better compare across different representation learning approaches, including XIRL, we use the L2 norm and the reward function in Equation~\eqref{eq:reward_form} for all representation learning methods including XTriplets.

\subsection{XIRL-Buckets}
The third method we study, takes advantage of the unsupervised nature of XIRL but also adapts the algorithm for a MQME dataset.
We propose XIRL-Bucket, which partitions the dataset into a number of ``buckets'' or bins based on ordinal labels provided by the user. 
We assume that a human labeller categorically assigns the trajectories into a bucket based on perceived performance. For example, the human could rate trajectories based on a 5-point Likert scale and then have a bucket for each ordinal rating 1 through 5.
We then train a representation by applying a TCC loss only amongst trajectories within the same bucket and use this representation as a reward function following the same procedure as XIRL.
Compared to X-RLHF, XPrefs, and XTriplets which require preference labels over a dataset, XIRL-Buckets only requires ordinal categorization which can be far less burdensome in terms of the number of times human queries. 



\section{EXPERIMENTS}


In this section, we seek to answer the following questions: (1) How does the quality of demonstrations degrade the learned reward and XIRL?
(2) Can we leverage different types of human feedback to learn good reward representations from MQME data?
(3) How do the different methods described above perform when learning from MQME data?


\subsection{Experimental Setup}

\paragraph{Domain} We conduct a series of experiments targeted at answering the aforementioned questions. For our experiments we use the X-MAGICAL imitation learning benchmark from \cite{zakka2022xirl}. An example of this task is shown in Figure~\ref{fig:qualitative-analysis}. The task involves pushing a set of blocks into the pink endzone using an agent of four possible embodiments: shortstick, mediumstick, longstick and gripper. The three stick embodiments all have the same action space but differ in their length, making the task easier for the longer embodiments and leading to qualitatively different optimal policies depending on the embodiment. Gripper not only has a different shape, but also has an extra action it can use to grip blocks with the pair of pincers on the agent.

\paragraph{MQME Data} To simulate a mixed-quality, mixed-embodiment dataset for our experiments, we took policies trained via RL on the ground-truth reward (number of blocks pushed to the goal divided by 3) for each of the embodiments listed above and then degraded these pretrained oracle policies by iteratively adding randomness to the action selection. This type of noise injection is inspired by prior work that found it resulted in diverse suboptimal behaviors~\citep{brown2020better,tien2022causal}. There are a total of 600 trajectories for each embodiment (200 training and 400 testing trajectories). The dataset is evenly partitioned based on the number of blocks that are pushed in by the agent.
By contrast, the X-MAGICAL dataset provided by \cite{zakka2022xirl} has the same embodiments as our MQME dataset but consists of approximately 1000 trajectories per embodiment (877 training and 98 testing trajectories), more than 4 times the amount of training data as our MQME dataset, all of which are exclusively successful demonstrations of the task.


We trained the reward model for X-RLHF using 5000 preference labels, all of which were obtained by sampling them from a larger set of procedurally generated preferences by comparing all the pairs of trajectories in our MQME dataset. The preferences were generated according to which trajectory in a pair of trajectories from the dataset has the higher average ground truth environment reward per step over the length of the trajectory. The reason the average reward per step was used is due to the longstick embodiment having a shorter time horizon for the task as it is significantly easier to complete the task with that particular embodiment. The synthetic preferences were meant to loosely mimic how a human would provide preferences observing the task. In a similar manner, we trained XTriplets for 4000 iterations with 32 triplets per batch where each triplet was formed by sampling with replacement from the MQME dataset and using the oracle synthetic labeler to order the triplets. We used much more training data than X-RLHF in an attempt to improve performance, but as we show later, we were not able to get performance comparable to X-RLHF even with the additional triplet feedback (see Appendix~\ref{app:hyperparams} for more details).

For XIRL-Buckets, we started with the same dataset of MQME 600 trajectories (200 for each training embodiment). We then simulated human ordinal ratings using the ground truth reward to partition the data into 18 buckets containing 32 trajectories each. This allowed us to pass an entire bucket into TCC as one batch, matching the batch size of~\cite{zakka2022xirl} for consistency across methods.



\subsection{Baselines}
In this section we describe the three baselines we compare against. All these baselines and methods use the same Soft Actor Critic code~\citep{zakka2022xirl} used in the original XIRL work:
(1)~\textbf{Reinforcement Learning on Ground Truth Reward.}
As an oracle, we run RL on the ground-truth reward from \cite{zakka2022xirl}.
At each step, the ground truth reward describes the fraction of total available blocks that are currently in the goal zone. 
(2)
\textbf{XIRL Trained on X-MAGICAL.} 
This method uses the full pipeline of XIRL as described in Section~\ref{sec:prelims} trained on the dataset of 200 successful demonstrations for each embodiment. This acts as an oracle since it provides the current-state-of-the-art performance for cross-embodiment IRL, but assumes access to near-optimal demonstrations for each embodiment. 
(3)
\textbf{XIRL Trained on Mixed Data.}
This method follows the XIRL pipeline but uses MQME data for TCC representation learning. The second step of the XIRL pipeline, goal embedding computation, is done with the same set of positive goal state examples we assume we have access to for the other methods studied in this paper. Thus, this baseline allows us to test the effect of mixed-quality data on XIRL and provides the main, non-oracle, baseline which we hope to significantly outperform.
(4)
\textbf{Goal Classifier}~\citep{vecerik2019practical}. We train a binary classifier where frames in the goal set ($G^*$) are positive examples and frames from the MQME dataset are negative examples. Following~\cite{zakka2022xirl}, the reward signal is the probability outputs of the model.


\subsection{Cross-Embodiment Learning from Mixed-Quality, Mixed-Embodiment Data}

\begin{wrapfigure}[26]{r}{0.58\textwidth}
  \centering    \includegraphics[width=\linewidth]{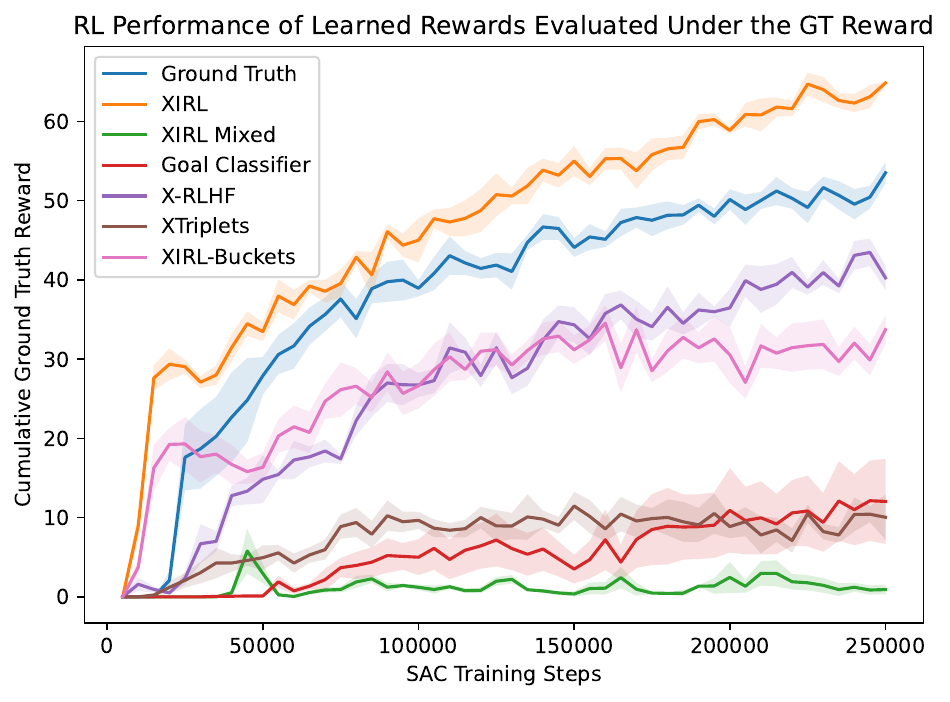}
    \caption{\textbf{Policy evaluations.} Shading denotes standard error bars. 
    Compared with the Ground Truth reward and XIRL (optimal demonstrations) oracles, we find that XIRL using mixed quality data suffers a significant performance drop, while X-RLHF and XIRL-Buckets perform much better. Simply training a goal classifier is insufficient and pretraining a representation based on triplet queries also fails to perform well.}
    \label{fig:final-comparison}
\end{wrapfigure}
To study how well our proposed approaches and baselines compare when evaluated on MQME data, all the reward models (except the oracle ground-truth RL baseline) were trained on 3 out of the 4 embodiments. We then evaluated the down-stream RL performance on the medium-stick held-out embodiment. To evaluate generalization performance, we took the average cumulative ground 
truth reward over 50 policy rollouts every 5000 RL training steps. These evaluation statistics were averaged over 5 seeds to account for randomness when learning a policy.


Figure \ref{fig:final-comparison} summarizes the results.
From this graph we can infer a few key findings of this experiment. Firstly, XIRL when trained on only successful trajectories, noticeably outperformed the other approaches by consistently pushing all three blocks in quickly and efficiently. Interestingly, XIRL actually outperforms the ground truth reward function by learning a good representation that successfully shapes the reward function, enabling efficient RL. On the other hand, XIRL Mixed, which was the same as XIRL, but trained on an MQME dataset, appears to completely collapse, unable to get a single block close to the goal zone. We hypothesize that this is because the mixed-quality data violates many of the strong assumptions that XIRL is founded on, in particular, the use of TCC to learn the latent video embedding.

Figure~\ref{fig:final-comparison} also shows that X-RLHF and XIRL-Buckets are both quite similar quantitatively with X-RLHF appearing to perform slightly better and XIRL-Buckets appearing to perform slightly worse. Interestingly, we find that XTriplets, despite training with more data than X-RLHF and XIRL-Buckets, fails to lead to good performance with performance similar to the goal classifier. This is surprising given the good performance reported in prior work. We hypothesize that the poor performance may be the result of the MQME data or the fact that we use an L2 norm distance metric rather than an optimal transport distance metric used in prior work~\citep{tian2023matters}. Future work should investigate these differences further and examine the performance of all of the studied methods when using different distance metrics such as optimal transport~\citep{cuturi2013sinkhorn}.
We conclude that X-RLHF and XIRL-Buckets successfully leverage mixed-quality, mixed-embodiment data, achieving much better performance than the prior state-of-the-art XIRL, when also evaluated on this data. However, there still is a noticeable gap between the XIRL on near-optimal data and the MQME methods. 
Thus, while mixed-quality data can be used to successfully transfer a policy to a new embodiment, it also appears to somewhat limit performance when compared with the performance achieved when training on near-optimal data. 
There are further insights that can be gained by quantitatively and qualitatively analyzing the learned rewards from these algorithms which will be discussed in the following sections. 

It is also important to note that quantitatively and qualitatively, the human effort differs across these methods. The human effort required for XIRL on X-MAGICAL for example, assumes perfect or optimal demonstrations which can be burdensome to obtain. X-RLHF on the other hand, is trained on 5000 preference labels. The number of times a human would be queried for XIRL-Buckets however, would be far fewer, equal to the total size of the dataset: approximately only 600 ordinal categorizations. The fact that XIRL-Buckets performs well despite using an order of magnitude fewer human labels is promising and future work should explore more of the design space for XIRL-Buckets. Finally, we note that XTriplets requires triplet preference queries which are likely more burdensome than preference queries. We leave it as an interesting area of future work to study the human factors involved in these different labeling schemes. 

\subsection{Reward Accuracy Analysis}\label{sec:reward_analysis}
\begin{table}[]
\small
\centering
\begin{tabular}{lcccccc}
\hline
              & \multicolumn{1}{l}{XIRL} & \multicolumn{1}{l}{XIRL Mixed} & \multicolumn{1}{l}{Goal Classifier} & \multicolumn{1}{l}{X-RLHF} & \multicolumn{1}{l}{XTriplets} & \multicolumn{1}{l}{XIRL-Buckets} \\ \hline
Kendall's Tau & 0.78                     & 0.61                           & 0.61                                & 0.80                       & 0.44                          & 0.52                             \\
Pairwise Acc. & 0.84                     & 0.75                           & 0.76                                & 0.85                       & 0.67                          & 0.71                             \\ \hline
\end{tabular}
\caption{\textbf{Accuracy Measures for Reward Learning}. Six reward representation methods are evaluated using a withheld set of 400 mediumstick mixed-quality demonstrations. Kendall's Tau measures the ordinal alignment between the cumulative ground truth rewards (oracle) and the cumulative learned rewards for each pair of demonstrations. Pairwise accuracy evaluates all ${400}\choose{2}$ pairs of trajectories and assigns each pair a score in $\{0, 1\}$ based on the learned reward alignment with the ground truth reward.}
\label{tab:reward_table}
\end{table}

To analyze the performance of each reward model, we evaluate the alignment of reward model outputs with the ordinal ground truth rankings of a withheld demonstration set. For each embodiment, we evaluate 400 test demonstrations of mixed quality. Each trajectory is then represented as the ordered pair $(r, \hat{r})$ representing the trajectory's cumulative ground truth reward ($r$) and the cumulative predicted reward ($\hat{r}$). The results of the withheld mediumstick embodiment for two accuracy metrics are shown in Table~\ref{tab:reward_table}. For brevity in the main body, we include correlation plots and performance for all reward learning embodiments in Appendix~\ref{app:reward_correlation}.

First, we use Kendall Rank Correlation Coefficient (also known as Kendall's Tau)~\citep{abdi2007kendall}, a statistical measure of determining correlation between two measured quantities. In our case, we wish to measure the alignment between $r$ and $\hat{r}$ for all $400 \choose 2$ demonstration pairs. Measurements for Kendall's Tau lie within the range $[-1, 1]$, where the agreement between the two quantities is in perfect agreement at 1, and perfect disagreement at -1. Second, using the same set of demonstration pairs, we evaluate the pairwise accuracy of the pairs by scoring a pair of trajectories (($r_i, \hat{r}_i$), ($r_j, \hat{r}_j$))  as "correct" if the ordering of $r_i, r_j$ is the same as the ordering of $\hat{r}_i, \hat{r}_j$ and "incorrect" otherwise.

Interestingly, we find that higher alignment scores between the cumulative estimated reward and the cumulative ground truth reward does not necessarily lead to better RL performance during policy training, a phenomenon reported in other work on preference learning~\citep{tien2022causal}. Both Goal Classifier and XIRL trained on MQME data score better on both reward learning metrics when compared with XIRL-Buckets and XTriplets, yet the latter approaches far outperform the former methods during RL training (Figure \ref{fig:final-comparison}). Furthermore, our results indicate that X-RLHF has the best Kendall correlation and accuracy on the validation set, surpassing even XIRL which was trained on X-Magical (near-optimal demonstrations). These findings suggest that quantifying the alignment between rewards is not necessarily a good indicator of down-stream RL performance. For example, such metrics could indicate a strong correlation between learned and oracle rewards even if a small and critical region of the state space is misidentified during training. Instead, in the following section, we examine more carefully the qualitative indicators of RL success.


\begin{figure*}[t!]
    \centering
    \begin{subfigure}{0.49\textwidth}
        \centering
        \includegraphics[width=\linewidth]{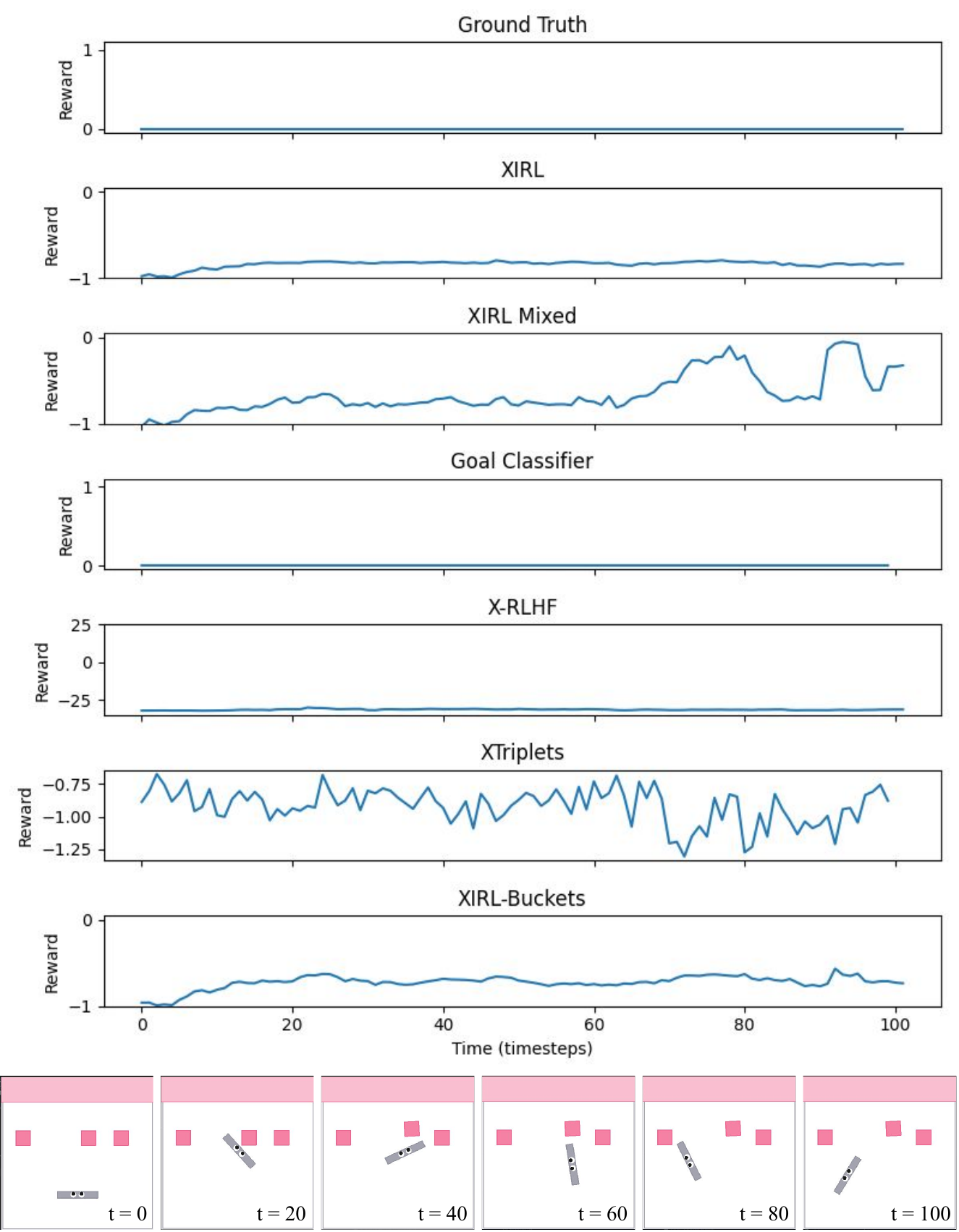}
        \caption{Learned Reward (Failed Trajectory)}
    \end{subfigure}%
    ~ 
    \begin{subfigure}{0.49\textwidth}
        \centering
        \includegraphics[width=\linewidth]{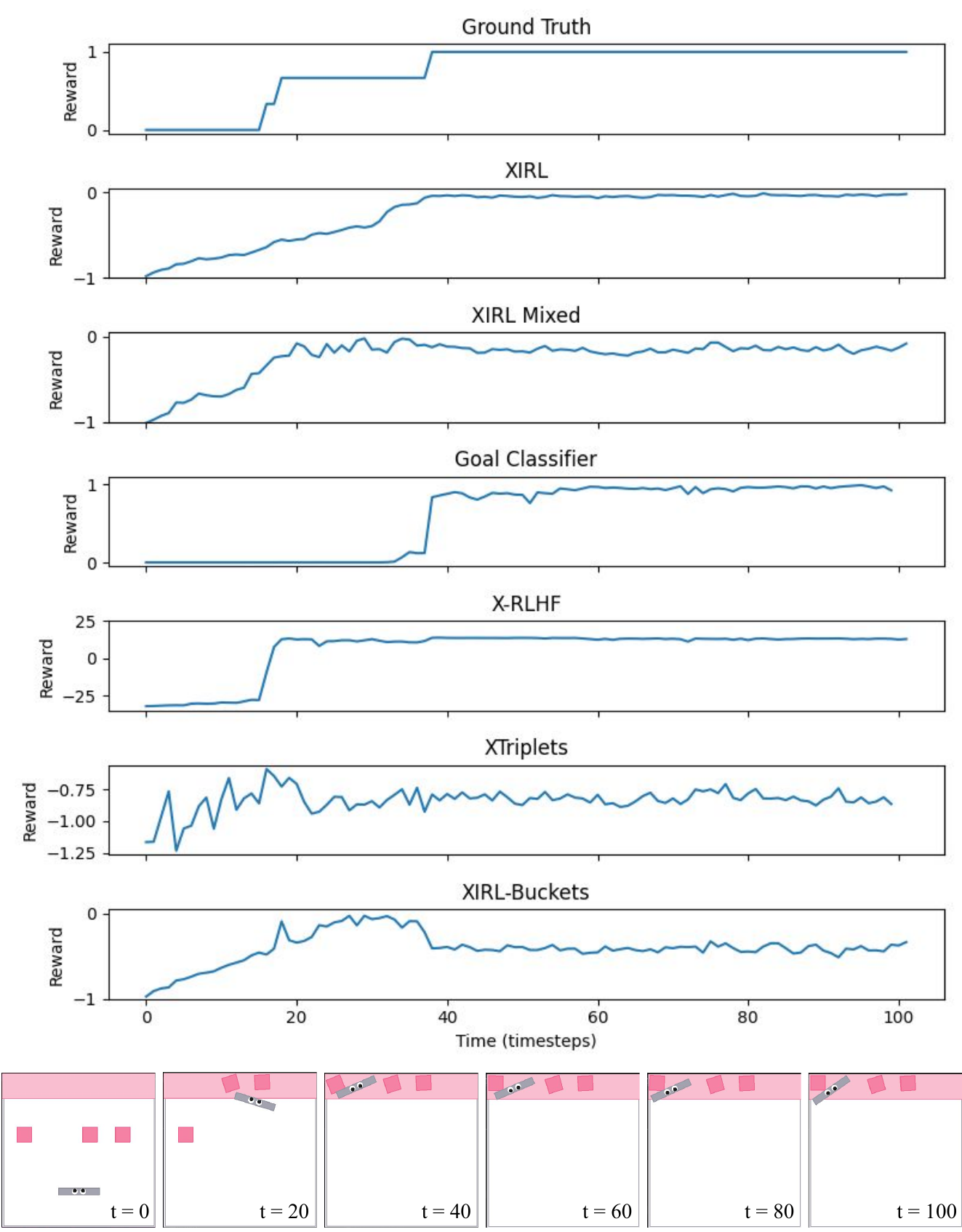}
        \caption{Learned Reward (Successful Trajectory)}
    \end{subfigure}
    \caption{\textbf{Qualitative Analysis of Learned Rewards and Representations}. We compare the true reward with the predicted rewards across a failed \textbf{(a)} and successful \textbf{(b)} trajectory.}
    \label{fig:qualitative-analysis}
\end{figure*}

\subsection{Qualitative Analysis of Learned Representations and Rewards}
We next performed a qualitative analysis of the learned representations and reward functions for the different methods.
Figure \ref{fig:qualitative-analysis} depicts the learned reward plotted against the timestep over the course of both a failed trajectory and a successful trajectory. This figure helps us visualize what reward signals and representations different approaches are learning.

Observing the shape of the learned reward for the unsuccessful trajectory, we can see that every learned reward except for XIRL trained on MQME has learned to associate actions that do not result in blocks being pushed with a low and near constant reward. The shaped reward for the successful demonstrations proves to be more informative with what these models are learning. 
XIRL trained on X-MAGICAL has the most shaped and dense reward, giving positive signals to the agent consistently throughout the task, even when it is moving away from the goal-zone to head back and gather more blocks. Interestingly, even XIRL on MQME has a surprisingly dense and informative signal, however this does not translate to final policy performance as it had the lowest average cumulative reward across all models. This can be explained by the fact that XIRL (MQME) likely assigns all end states as having high reward since the TCC alignment across mixed demonstrations will push the end frames of the videos are aligned to be considered successful. We can see evidence of this in the curve for XIRL (MQME) for the failed trajectory, where the predicted reward still spikes upward. Finally, XTriplets poorly predicts the reward for the failed trajectory and incorrectly attributes relatively high reward to states that do not progress the task, which is supported by XTriplets' poor RL performance.

X-RLHF appears to learn a similarly shaped reward to that of the ground truth reward (a step function 
 representing the fraction of total available blocks that are currently in the goal zone). It gives clear spikes in reward signals when blocks are pushed in, but fail to provide information in between these key states. 
 On closer inspection of the policies learned by the X-RLHF reward
 we found they learned a greedy strategy to get as many blocks as possible into the goal zone with a single push. 
 The learned models do not appear to associate a strong reward signal for actions that occur after getting 2 blocks in. Leveraging online queries to refine the reward representation could overcome this problem and is an interesting area for future work.

\section{CONCLUSION}
In this paper we introduce the novel problem setting of cross-embodiment learning from mixed-quality data.
Collecting near-optimal demonstrations in complex environments is challenging and human demonstrations are often noisy or suboptimal.
We propose and evaluate X-RLHF, XTriplets, and XIRL-Buckets as three potential algorithms to help address these shortcomings. 
Our empirical results demonstrate that, XIRL~\citep{zakka2022xirl}, the prior state-of-the-art approach to cross-embodiment IRL suffers a large degradation in performance when not all demonstrations are near-optimal. By contrast, X-RLHF and XIRL-Buckets both showcase the ability to leverage human feedback over mixed-quality, mixed-embodiment data to learn a reward function that is embodiment independent and enables the ability to generalize this reward to out-of-distribution embodiments. 
An exciting area of future work is to explore a combination of some of our proposed methods. For example, X-RLHF and XIRL-Buckets appear to have somewhat complimentary reward shapes, could a linear combination of the two lead to a more informative reward? Similarly, we could leverage our goal embedding calculation and training as a finetuning step after a run of a TCC algorithm. Another area of future work is to study the human factors involved in different forms of human feedback as they relate to representation alignment and cross-embodiment learning.  Finally, using active preference learning could enable more label-efficient algorithms~\citep{biyik2018batch,wilde2020active,shin2023benchmarks}.
\bibliography{main}
\bibliographystyle{rlc}

\newpage
\appendix

\section{Further Analysis of XPrefs}\label{app:static-vs-dynamic}

\begin{figure*}[t!]
    \centering
    \begin{subfigure}{0.48\textwidth}
        \centering
        \includegraphics[width=\linewidth]{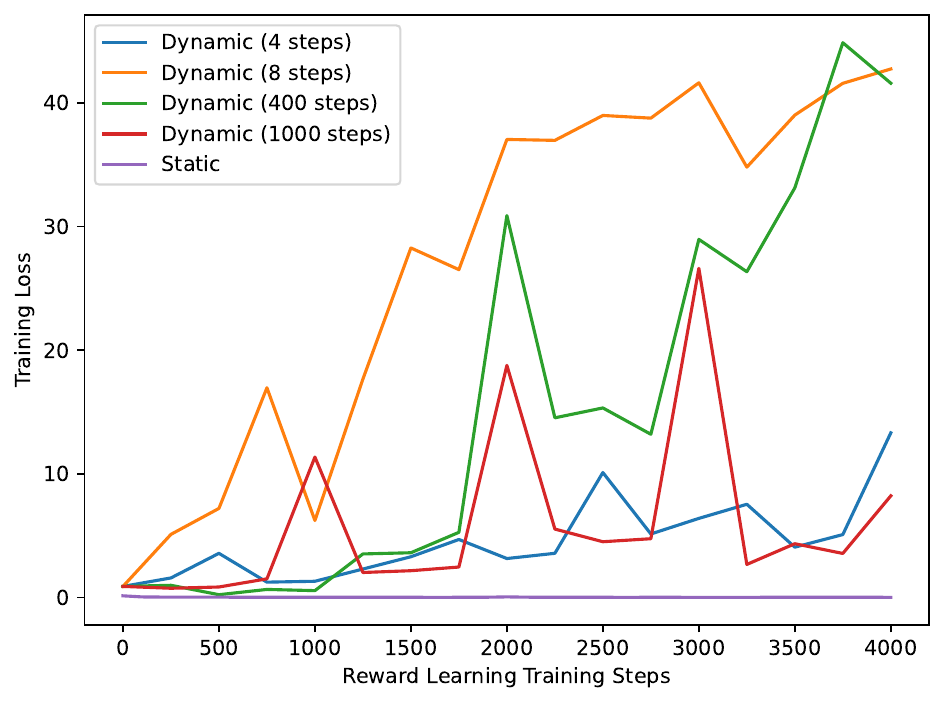}
        \caption{Static vs. Dynamic Goal Embeddings}
        \label{fig: static-dynamic}
    \end{subfigure}%
    ~ 
    \begin{subfigure}{0.50\textwidth}
        \centering
        \includegraphics[width=\linewidth]{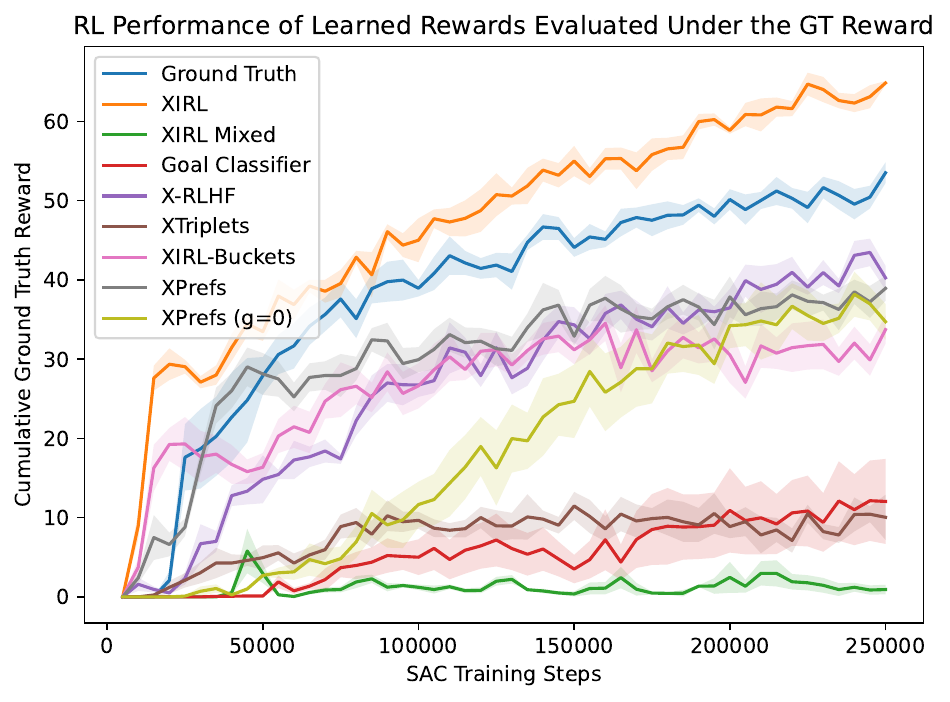}
        \caption{RL Performance}
        \label{fig: xprefs-rl-performance}
    \end{subfigure}
    \caption{\textbf{XPrefs Performance.} \textbf{(a)} Static vs. Dynamic Goal Embeddings for XPrefs. We find that using dynamic goal embeddings that were periodically updated every X steps during training leads to training instabilities, hindering good representation learning. By contrast, using a static goal embedding leads to a convergent training loss. \textbf{(b)} The RL performance of XPrefs and XPrefs with the goal representation ($g$) hard-coded as the origin. In both cases, XPrefs obtains very similar results to X-RLHF}
\end{figure*}

One of the design choices mentioned in the section describing Xprefs, is whether to use a static or dynamic goal embedding. At first glance, a dynamic embedding seemed to be the best way to ensure the guarantee of a global minima so we performed a preliminary experiment with varying frequencies of goal embedding updates in the training process. Figure \ref{fig: static-dynamic} showcases the training losses over training steps we observed for the various frequencies of updating the goal embedding $g$ described in the XIRL preliminaries. We experimented with frequencies of 4, 8, 400 and 1000 and compared the results with using static goal embedding frozen before training. Our results provide evidence that periodically updating the goal embedding causes instabilities since it induces a moving target during learning.

In Figure~\ref{fig: static-dynamic}, the loss curve when training XPrefs and updating the goal embedding every 1000 steps is particularly interesting as there is a clear spike in loss every time the goal embedding is updating, leading to an intuition that the updates lend to an instability and make it more difficult to settle in a local minimum. Our results provide evidence that periodically updating the goal embedding causes instabilities since it induces a moving target during learning. 
We settled on using a static embedding model which resulted in a much more stable and meaningful learned reward, as shown by the clear convergence of the loss in Figure~\ref{fig: static-dynamic} when using a static goal embedding. We theorize that fixing an arbitrary point in the embedding space as our goal state is appropriate and in fact beneficial to learning an embodiment-independent state representation as it is akin to fixing an origin for the space and fitting the state distribution around it in a way that is locally optimal.

In section~\ref{sec:representations_from_prefs}, we theorized that XPrefs is simply X-RLHF with a non-positive reward function. To emperically support this notion, we trained two reward models and compared RL performance to X-RLHF (Figure~\ref{fig: xprefs-rl-performance}). The first model is XPrefs with a static goal embedding that utilizes the goal set ($g^*$), as described in the previous paragraph. The second model (notated "Xprefs (g=0)" in Figure~\ref{fig: xprefs-rl-performance}), tests our hypothesis that the goal embedding is an arbitrary bias that does not help reward learning. As predicted, the performance of both XPrefs methods does not show any improvement to RL training compared to X-RLHF. Therefore, there is no empirical evidence to support that XPrefs with static goal embeddings leads to improved cross-embodiment learning.



\section{Reward Correlation Analysis}\label{app:reward_correlation}

\begin{table}[h]
\centering
\begin{tabular}{llcccccc}
              &                          & \multicolumn{1}{l}{XIRL} & \multicolumn{1}{l}{XIRL Mixed} & \multicolumn{1}{l}{Goal Classifier} & \multicolumn{1}{l}{X-RLHF} & \multicolumn{1}{l}{XTriplets} & \multicolumn{1}{l}{XIRL-Buckets} \\ \hline
\multicolumn{2}{l}{\textbf{Gripper}}     & \multicolumn{1}{l}{}     & \multicolumn{1}{l}{}           & \multicolumn{1}{l}{}                & \multicolumn{1}{l}{}       & \multicolumn{1}{l}{}          & \multicolumn{1}{l}{}             \\
              & Kendall's Tau            & 0.72                     & 0.57                           & 0.46                                & 0.79                       & 0.54                          & 0.58                             \\
              & Pair Accuracy            & 0.79                     & 0.72                           & 0.67                                & 0.83                       & 0.71                          & 0.73                             \\ \hline
\multicolumn{2}{l}{\textbf{Longstick}}   & \multicolumn{1}{l}{}     & \multicolumn{1}{l}{}           & \multicolumn{1}{l}{\textbf{}}       & \multicolumn{1}{l}{}       & \multicolumn{1}{l}{}          & \multicolumn{1}{l}{}             \\
              & Kendall's Tau            & 0.81                     & 0.69                           & 0.75                                & 0.81                       & 0.65                          & 0.53                             \\
              & Pair Accuracy            & 0.83                     & 0.77                           & 0.80                                & 0.83                       & 0.75                          & 0.69                             \\ \hline
\multicolumn{2}{l}{\textbf{Shortstick}}  & \multicolumn{1}{l}{}     & \multicolumn{1}{l}{}           & \multicolumn{1}{l}{\textbf{}}       & \multicolumn{1}{l}{}       & \multicolumn{1}{l}{}          & \multicolumn{1}{l}{}             \\
              & Kendall's Tau            & 0.76                     & 0.57                           & 0.47                                & 0.80                       & 0.22                          & 0.60                             \\
              & Pair Accuracy            & 0.82                     & 0.72                           & 0.68                                & 0.84                       & 0.55                          & 0.74                             \\ \hline
\multicolumn{2}{l}{\textbf{Mediumstick}} & \multicolumn{1}{l}{}     & \multicolumn{1}{l}{}           & \multicolumn{1}{l}{\textbf{}}       & \multicolumn{1}{l}{}       & \multicolumn{1}{l}{}          & \multicolumn{1}{l}{}             \\
              & Kendall's Tau            & 0.78                     & 0.61                           & 0.61                                & 0.80                       & 0.44                          & 0.52                             \\
              & Pair Accuracy            & 0.84                     & 0.75                           & 0.76                                & 0.85                       & 0.67                          & 0.71                            
\end{tabular}
\caption{\textbf{Accuracy Measures for Reward Learning (Full Table)}. An expanded version of Table \ref{tab:reward_table} that includes quantitative reward metrics for all 4 embodiments. Reward learning is trained on demonstrations from the gripper, longstick, and shortstick embodiments. The withheld mediumstick embodiment is used to evaluate whether the learned reward correctly generalizes to new embodiments.}
\label{tab:appendix_reward_table}
\end{table}

\begin{figure}
    \centering
    \includegraphics[width=1\linewidth]{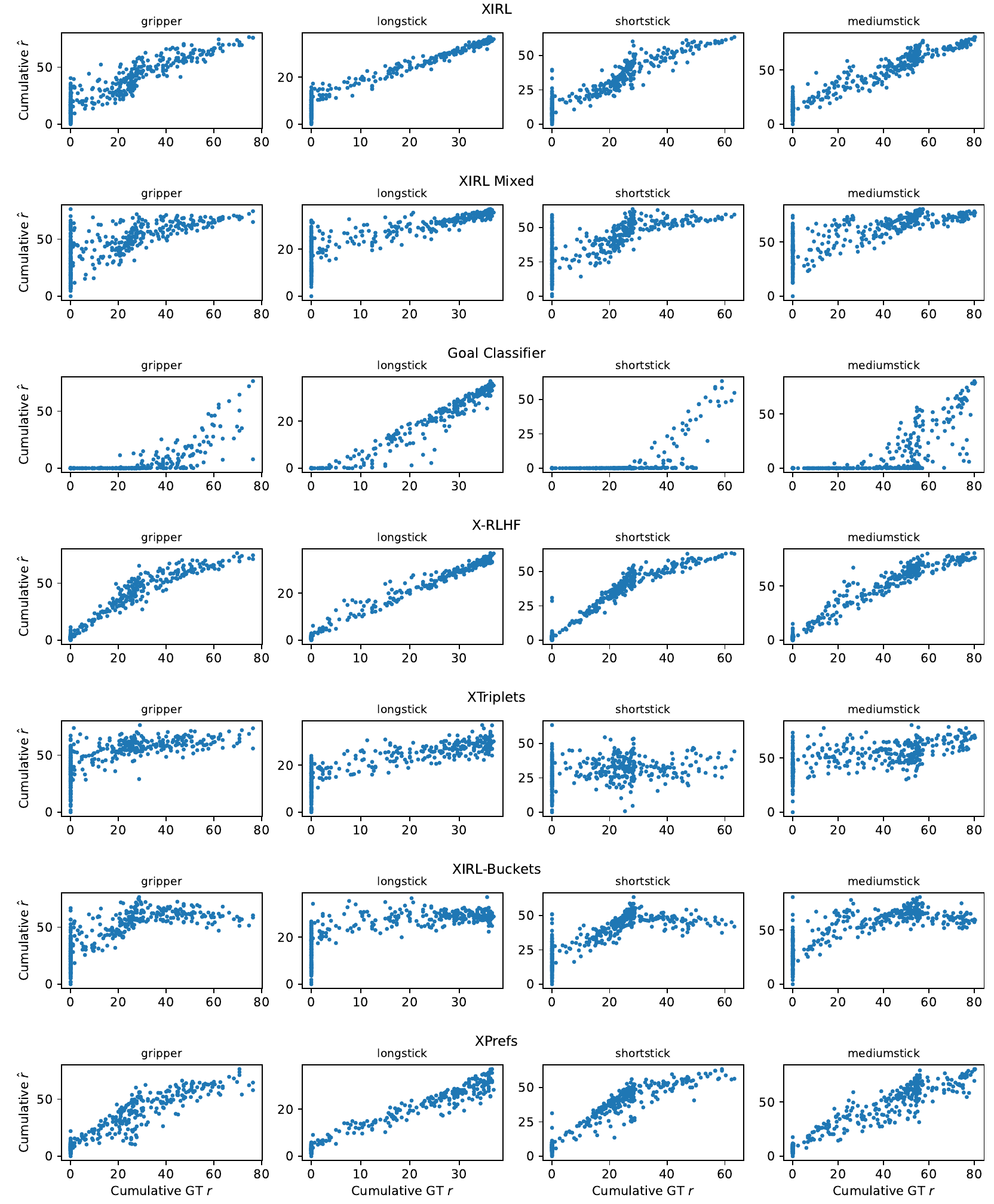}
    \caption{\textbf{Correlation Plots.} The correlation between ground truth reward ($r$) and learned reward ($\hat{r}$, normalized to range of $r$) for the reward learning test demonstration set. Rewards shown are cumulative rewards over full trajectories. Reward learning trains on the XMagical gripper, longstick, and shortstick embodiments (Columns 1-3) and then is tested on the withheld mediumstick embodiment (Column 4).}
    \label{fig:reward_correlation}
\end{figure}

In addition to analyzing the reward accuracy for the withheld mediumstick embodiment (Section \ref{sec:reward_analysis}), we also evaluate the quality of the learned reward on the validation sets for the in-distribution embodiments. Table~\ref{tab:appendix_reward_table} contains the Kendall's Tau~\citep{abdi2007kendall} and Pairwise Accuracy scores for all 4 embodiments and all evaluated methods.

In Figure~\ref{fig:reward_correlation}, we include the scatter plots for the data used to calculate these metrics by plotting each trajectory's cumulative ground truth reward ($r$) and cumulative learned reward ($\hat{r}$).

Notably, despite withholding  mediumstick data at training time, Table~\ref{tab:appendix_reward_table} indicates that there is almost no degradation in performance when comparing the testing demonstrations for the in-distribution embodiments compared to the testing demonstrations for the out-of-distribution embodiment. Given mixed embodiment data, we would expect this to be the case as our reward models should not be overfitting to the features of any one agent model. 

The correlation plots are also a helpful visualization for understanding how well reward learning correlates with the ground truth reward, which is the quantity we used to synthetically generate preferences. For each emoboiment, we would expect to see a positive linear correlation between the ground truth (x-axis) and learned reward (y-axis). In almost all methods and embodiments, it is clear that some notion of correct representation was encoded at training time as shown in the largely positive trends shown in each plot. 

Many of the methods have a wide distribution of points at x=0. This is indicative of a learned reward signal that differentiates between states that have zero ground truth reward. Because the ground truth reward function is sparse (reward signal only increases after a block is pushed to the goal region), therefore several models have learned something more dense than the ground truth reward. If the reward model learned incorrectly to attribute high reward to poor trajectories, as we suspect is the case for XIRL Mixed, this almost certainly has a detrimental effect on downstream RL. As shown in XIRL, some distribution at x=0 is likely helping the performance of the policy, rather than hurting it, as the XIRL policy outperforms the Ground Truth reward during RL.

\section{Experiment Hyperparameters}\label{app:hyperparams}
For reproducibility, we include the technical details of experimentation and model training below. For access to our source code, we refer you to our project website at \url{https://sites.google.com/view/cross-irl-mqme/home}.

\subsection{Reward Learning}
All reward learning uses a pretrained Resnet18 with a linear projection head to embed 3x112x122 images into either a 32 dimensional latent space (XIRL, XIRL Mixed, XTriplets, XPrefs, XIRL-Buckets) or a single real-valued output (X-RLHF, Goal Classifier). All networks use a batch size of 32 and use the Euclidean (L2) distance as the embedding similarity metric.

XIRL, XIRL Mixed, and XIRL-Buckets all use the TCC Loss with the exact same configuration as~\cite{zakka2022xirl}. Goal Classifier uses BCELossWithLogits Loss and XTriplets, XPrefs, and X-RLHF use a Cross Entropy Loss. All networks use the Adam optimizer with learning rate 1$e$-5.

We trained XTriplets for 4000 iterations with 32 triplets per batch where each triplet was formed by sampling with replacement from a dataset of 600 offline video trajectories. We allow for many more triplets than the 5,000 pairwise preferences that X-RLHF is allotted and still observe poor performance from XTriplets. 

\subsection{Reinforcement Learning (Soft Actor-Critic)}
All of our learned representation methods use the exact same RL parameters to ensure comparability. Our RL parameters are consistent with the original XIRL work~\citep{zakka2022xirl}. We train RL for 250,000 training steps, where the first 5,000 steps are randomly taken to populate a replay buffer with capacity $1e6$. Performance is measured by evaluating 50 episodes every 5,000 steps. The actor and critic both share the same MLP architecture with 2 hidden layers of 1024 nodes each. We use $\gamma = 0.99$ for the discount factor, 1$e$-4 for the learning rate of both actor and critic, and a batch size of 1024.





\end{document}